\documentclass[amt, manuscript]{copernicus}

\graphicspath{{figures/}}

\begin{document}

\nolinenumbers

\title{Flight Contrail Segmentation via Augmented Transfer Learning with Novel SR Loss Function in Hough Space}

\Author[1]{Junzi}{Sun}
\Author[1]{Esther}{Roosenbrand}

\affil[1]{Faculty of Aerospace Engineering, Delft University of Technology,Kluyverweg 1, 2629 HS, Delft, the Netherlands}

\correspondence{Junzi Sun (j.sun-1@tudelft.nl)}

\runningtitle{TEXT}

\runningauthor{TEXT}

\received{}
\pubdiscuss{} 
\revised{}
\accepted{}
\published{}

\firstpage{1}

\maketitle

\begin{abstract}
  Air transport poses significant environmental challenges, particularly regarding the role of flight contrails in climate change due to their potential global warming impact. Traditional computer vision techniques struggle under varying remote sensing image conditions, and conventional machine learning approaches using convolutional neural networks are limited by the scarcity of hand-labeled contrail datasets. To address these issues, we employ few-shot transfer learning to introduce an innovative approach for accurate contrail segmentation with minimal labeled data. Our methodology leverages backbone segmentation models pre-trained on extensive image datasets and fine-tuned using an augmented contrail-specific dataset. We also introduce a novel loss function, termed SR Loss, which enhances contrail line detection by transforming the image space into Hough space. This transformation results in a significant performance improvement over generic image segmentation loss functions. Our approach offers a robust solution to the challenges posed by limited labeled data and significantly advances the state of contrail detection models.
\end{abstract}


\section{Introduction}

Air transport is critical to the global economy but presents significant environmental challenges like other transports relies on fossil fuel. They are emit greenhouse gases like carbon dioxide and nitrous oxide. However, unlike other types of transports, aircraft-generated contrails could cause impact the climate more due to their potential global warming contribution \citep{grewe2017mitigating}. Effective monitoring and understanding of contrails are therefore essential to manage air transport's climate impacts.

\subsection{Computer vision approaches}

Contrail detection traditionally utilizes computer vision tasks, given that it involves identifying linear features. Advancements in computational power have led to early research by \cite{weiss1998automatic} and \cite{mannstein1999operational} introducing methods to detect contrails in satellite images using various image processing techniques, such as ridge classification, Hough transform, and contrail line segmentation with linear filters.

In later research by \cite{vazquez2010automatic}, methods have been extender with new algorithms to track contrails as they age, drift, and spread. This algorithm tests for contrails in an image, computes contrail masks considering the surrounding pixels, and ultimately determines the contrail cluster's overall shape, facilitating contrail evolution studies. In another approach, \cite{zhang2017verification} combined visual contrail detection with meteorological data, leading to a contrail occurrence and persistence index that enables studying contrail coverage.

With advanced image processing algorithms, \cite{minnis2013linear} investigated linear contrails and contrail cirrus clouds' properties using a blend of remote sensing imagery processing techniques. The study utilized a combination of these approaches, including brightness temperature differences \citep{ackerman1996global}, infrared bispectral technique \citep{bedka2013properties}, and visible infrared shortwave-infrared split-window technique \citep{minnis2011ceres}.

\subsection{Machine learning approaches}
Over the past a few years, machine learning, especially supervised machine learning, has been adopted widely in remote sensing research. Thus, research have been applying machine learning for the contrail detection tasks. \cite{kulik2019satellite} utilized an autoencoder-based convolutional neural network model to identify contrails in satellite images. However, despite its success in contrail detection, the model could not determine their exact location due to the simplicity of the employed machine learning model. A parallel approach by \cite{siddiqui2020atmospheric} also used a convolutional neural network, focusing on whether contrails appear in a frame captured by a ground-based camera.

A recent research by \cite{mccloskey2021human} has provided a modest set of human-labeled landsat images for the research community. Another recent study by \cite{ng2023opencontrails}, in a more comprehensive effort, aimed to construct an open dataset for contrails over the United States, using the GOES-16 satellite imagery. In the course of assembling this dataset, a convolutional network was employed. This network was designed to take a series of temporally sequenced images as inputs and subsequently detect and outline contrail segments. The research show promising results, however, the training of such model requires very high performance hardwares and the details of the models are not made available by the paper.

\subsection{Research gaps}

It has been found that traditional computer vision-based approaches have shown limitations in detecting contrails due to the complexity of satellite images captured under varying conditions. Earlier machine learning approaches often employed simpler convolutional neural network models primarily used to determine an image's contrail presence, often requiring a large volume of labeled data.

Moreover, current literature lacks research specifically tailored to machine learning approaches for contrail detection, distinguishing it from other image segmentation or detection tasks. In particular, there are no adequate loss functions optimized for linear features like contrails to be utilized in training. This shortcoming makes detection particularly challenging at lower resolutions, especially when multiple contrails are in close proximity. 

\subsection{Contribution of this study}

This paper aims to present a machine learning model that merges transfer learning using pre-trained backbone models with augmented satellite images to improve the training efficiency. This approach enables the effective training of contrail detection models with a minimal dataset on standard computer hardware. 

Firstly, we complement the transfer learning with data augmentation methods that manipulate both the input images and the labeled contrails. This creates a random input for each training epoch, thereby equipping the neural network to manage images with varying properties, including contrasts, gamma corrections, and perspectives.

Secondly, an major contribution is the customized loss function, SR Loss, that is specifically designed for detecting and segmentation of contrails, taking advantage of the linear shapes of contrails.

We also offer open access to the source code, imagery data, and the neural network model. Moreover, we develope and share the models implemented in both PyTorch and TensorFlow, and we perform thorough evaluations of the performances of our models openly.

The remainder of the paper is as follows. Section \ref{sec:method} explains the data, neural network model, and augmentation of imagery data. Section \ref{sec:experiment} presents the model's experiments and results. Finally, Section \ref{sec:discussions} and Section \ref{sec:conclusion} offer the discussions and conclusion of this research.

\section{Data processing} \label{sec:method}

\subsection{GOES data}
Contrails and cirrus clouds share atmospheric similarities, allowing the utilization of infrared channels commonly employed for cirrus cloud identification to also detect contrails. To isolate the presence of optically thin cirrus clouds and eliminate background and ground interference, a pre-process technique involving the calculation of brightness temperature differences (BTD) is employed in this study.

\begin{figure}[hpt!]
  \includegraphics[width=0.96\columnwidth]{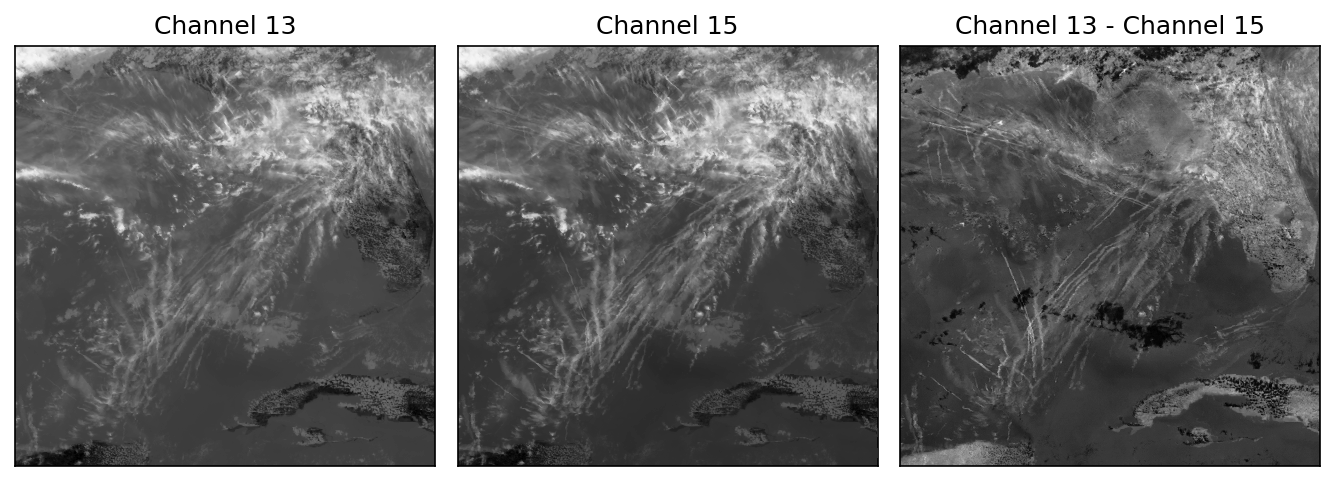}
  \centering
  \caption{Images from channel 13 (12.3 µm) and 15 (10.35 µm) for GOES-16 satellite over the Gulf of Mexico. The brightness temperature differences (BTD) of the two channels are shown in the last image.}
  \label{fig:preprocess_goes_image}
  \end{figure}

This technique subtracts one infrared channel from another to generate BTD images. In our case, specifically, difference between 12.3 µm and 10.35 µm for the GOES-16 satellite is obtained. This is demonstrated in Figure \ref{fig:preprocess_goes_image}, which showcases an resulting of a BTD image featuring contrails.

After days and regions are identified with contrail occurrence, the GOES-16 data is downloaded using the \emph{goes2go} package. These files are processed with the \emph{netCDF4} library, where channel 13 (10.35 µm) is subtracted from channel 15 (12.3 µm) to produce the final image used for training and labelling. 

As a geostationary satellite, GOES-16 data use its own projection format to ensure full-earth disk coverage. We then convert the image into a local project using the \emph{pyproj} Python libaray.

\subsection{Contrail labeling}

\begin{figure}[ht!]
\centering
\includegraphics[width=0.31\columnwidth]{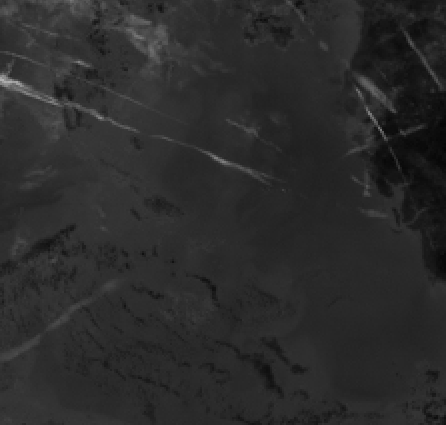}
\includegraphics[width=0.31\columnwidth]{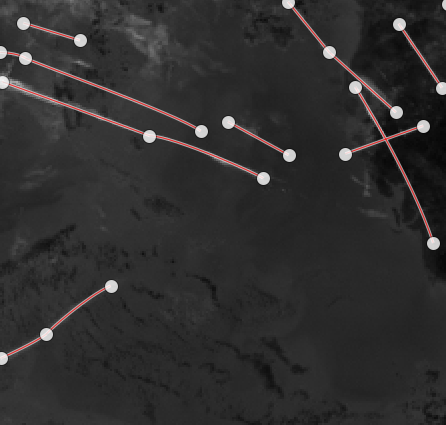}
\includegraphics[width=0.31\columnwidth]{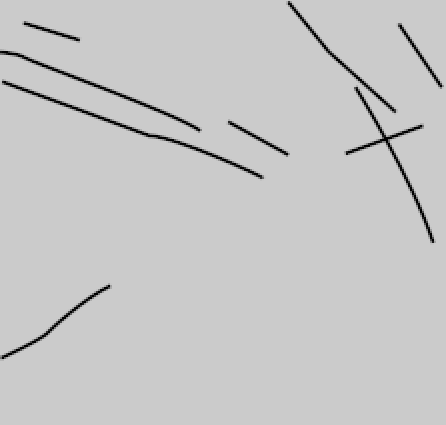}
\caption{The process of manual contrail labeling. From left to right: base image, overlay of labeled contrail paths, and final contrail mask image (illustrated with a gray background).}
\label{fig:labeling}
\end{figure}

The satellite images are processed with GIMP - the GNU Image Manipulation Program. Contrails are first traced with \emph{paths}. Based on all identified contrail paths, the mask image is generated with strokes of approximately two pixels on all paths. Figure \ref{fig:labeling} shows the process of generating the mask from the original image.

In total, around 30 images at two different locations, including San-Francisco and Florida, are selected and manually masked with contrails. Within these 30 images, 10 are reserved for evaluating the performance of the model, which are not use in training of the model.

\section{Segmentation model design}

\subsection{U-Net}

The U-Net model \citep{ronneberger2015u}, originally developed for biomedical image segmentation in 2015, has gained significant attention in segmentation and generation research studies due to its high accuracy and efficiency. It is chosen to be the underlying model for the contrail segmentation task in this paper.

\begin{figure}[ht!]
  \centering
  \includegraphics[width=0.95\columnwidth]{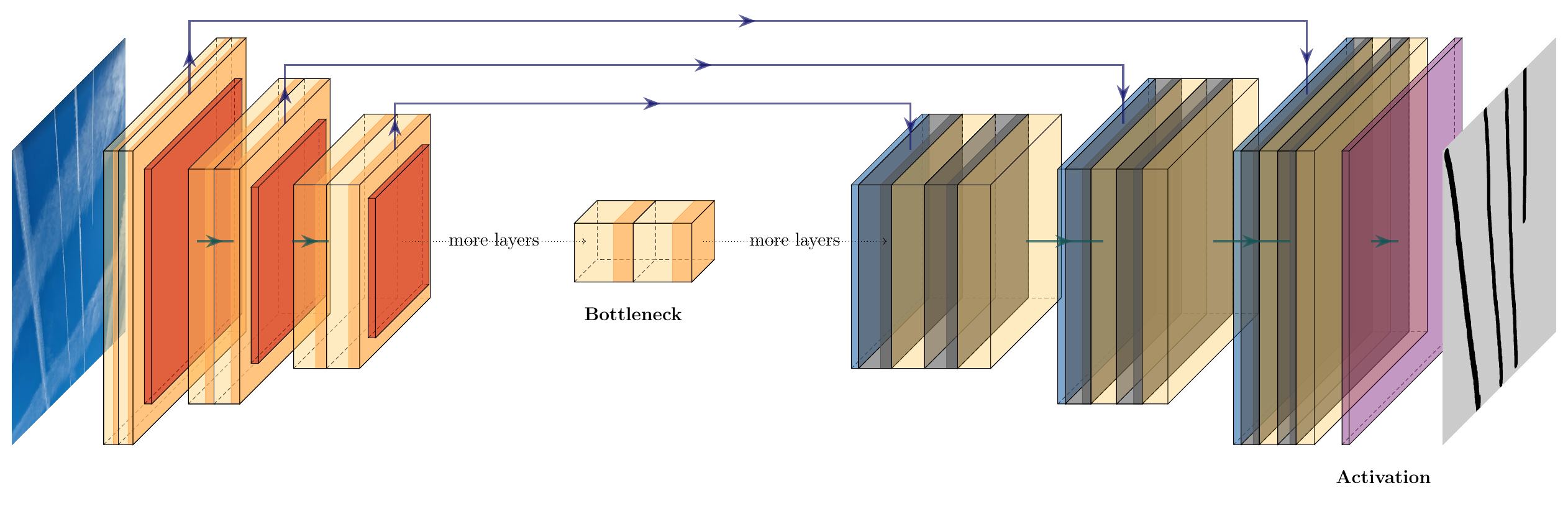}
  \caption{Illustration of the U-Net neural network}
  \label{fig:unet}
\end{figure}

The U-Net is a fully convolutional neural network (CNN) and consists of two main components: the contracting and expansive paths, which are also sometimes referred to as encoder and decoder paths. The overall network is illustrated in Figure \ref{fig:unet}.

The encoder or contracting path follows the traditional structure of a CNN, consisting of multiple convolutional and pooling layers. This path is designed for capturing the context and features of the input image. The encoder gradually reduces the spatial dimensions of the feature maps while increasing their depth. This process expects the abstracted semantics to be captured.

The decoder path of a U-Net generates a pixel-wise segmentation map that corresponds to the input image. It up-samples the layers using transposed convolutions (or de-convolutions). This path helps to re-construct spatial resolutions and thus lead to the generation of masks of contrail that correspond to the input image.

For each pair of corresponding encoder and decoder layers, a skip connection is established. The skip connection concatenates features from both layers and allows the network to combine both high-level and lower-level features for classification.

\subsection{ResNet}

\begin{figure}[ht!]
  \centering
  \includegraphics[height=0.35\columnwidth]{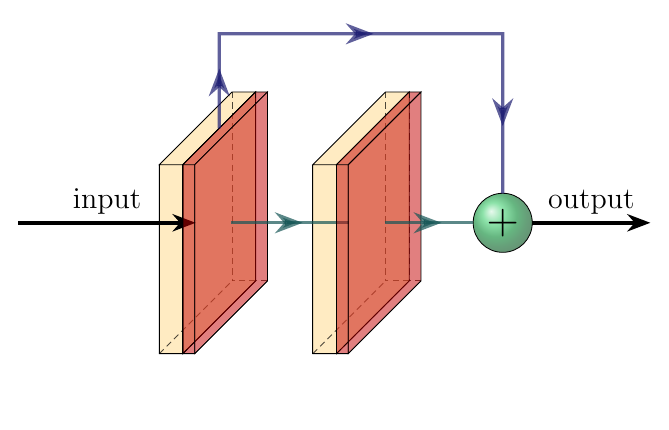}
  \caption{Illustration of a residual block in ResNet}
  \label{fig:resblock}
\end{figure}

In the segmentation task, U-Net is often combined with ResNet (also known as Residual Network) \citep{he2016deep}, a type of network designed to address the challenge of training deep neural networks, especially the diminishing gradients in back propagation. The residual block also employs skip connections in addition to the convolution layers. Instead of directly learning the relationship between input and output

\begin{equation} \label{eq:mapping}
  \text{output} = F (\text{input}),
\end{equation}

\noindent it learns the difference between the input and desired output

\begin{equation} \label{eq:res}
  \text{output} = \text{input} + F_\text{res}(\text{input}),
\end{equation}

\noindent where $F$ represents the direct mapping, and $F_\text{res}$ represents the residual mapping. The structure of the residual block in a ResNet can be seen in Figure~\ref{fig:resblock}

\subsection{Combining U-Net and ResNet}

\begin{figure}[ht!]
  \centering
  \includegraphics[width=0.95\columnwidth]{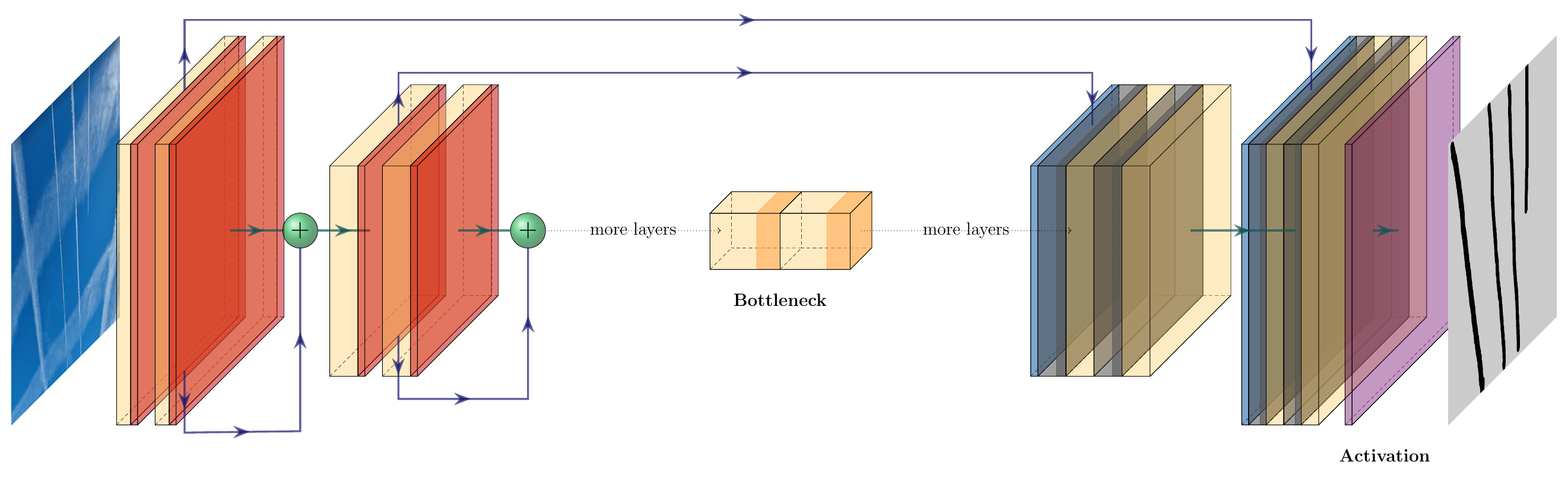}
  \caption{Illustration of the proposed ResUNet architecture}
  \label{fig:resunet}
\end{figure}

When U-Net and ResNet are combined into one model, the network can take advantage of both architectures. It inherits the ability to capture fine feature details from U-Net and the ability to learn deep representations from ResNet. We adopt a common approach, U-Net with ResNet encoder (ResUNet), where the ResNet acts as the backbone for feature extraction, and U-Net acts as the decoder for segmentation. In Figure \ref{fig:resunet}, the illustration of the overall structure of our segmentation model is shown.

\subsection{Conventional loss functions}

The choice of loss function is critical part for any machine learning task, as it determines what to optimize during the training. Two conventional loss functions are tested in the training of the neural network, which are Focal loss \citep{lin2017focal} and Dice loss \citep{sudre2017generalised}. We also invented a new loss function specially for the contrail detection problem, which is discussed in detail in Section \ref{sec:sr_loss}.

Here, both the Focal and Dice loss functions are relevant choices, as the classes (contrail and non-contrail) are highly imbalanced in our study, and these loss functions do not heavily penalize the prediction the majority class (non-contrail pixels).

Focal loss is designed to focus the learning process on misclassified and hard classifications. It applies a specific \emph{modulating} factor, $(1-p_t)^\gamma$, to the commonly used cross-entropy loss:

\begin{equation} \label{eq:focal_loss}
  \begin{split}
  L_\text{Focal}(p_t) &= -(1-p_t)^\gamma \log(p_t) \\
  p_t&=\begin{cases}
    p,  & \, \text{if} \quad y=1 \\
    1-p, & \, \text{otherwise}
  \end{cases}
\end{split}
\end{equation}

\noindent Where $\gamma (\gamma>0)$ is the scaling factor introduced by Focal loss. When $\gamma=1$, the loss is equivalent to binary cross-entropy loss. Increasing $\gamma$ will give more focus on less well-classified examples.

Dice Loss is a loss function that is commonly used in segmentation tasks. It is based on the Dice coefficient, which is a metric that measures the similarity of predicted classes and true classes:

\begin{equation} \label{eq:dice_loss}
  L_\text{Dice}(p, g) = 1 - \frac{2 g p + 1}{g + p + 1}
\end{equation}

\noindent where $p$ represents the probability of prediction for a pixel belonging to the target class (contrail or non-contrail), $g$ is the ground truth class for that pixel. The Dice loss is also often represented in the logarithmic form, as follows:

\begin{equation} \label{eq:dice_loss_log}
  L_\text{logDice}(p, g) = \log \left( \frac{2 g p + 1}{g + p + 1} \right)
\end{equation}

\section{SR Loss at Hough space to improve contrail segmentation} \label{sec:sr_loss}

The previous generic loss functions primarily help the neural network determining whether a pixel should be classified as a contrail based on differences with adjacent pixel areas via convolutions. And they do not explicitly considers the potential shapes of the contrail. 

In designing our new loss function, we aim to specifically take into account this inherent linear shape of contrails. To achieve this, we implement a transformation of the image into Hough space, where each point signifies a line in the image space. Throughout the training process, we will also minimize the differences in Hough space between the predicted contrail formations and the target. In this section, we will explain the design of this new loss function.

\subsection{Hough space and transformation}

Hough transformation is a technique proposed by Paul Hough in 1962 \citep{hough1962method} that has been widely adopted in image processing and computer vision. It is often employed to detect and extract linear features in images.

The Hough transformation first converts the common linear representation of

\begin{equation}
  y = a x + b
\end{equation}

\noindent into polar coordinate format

\begin{equation}
  \rho = x \cos \theta + y \sin \theta
\end{equation}

\noindent where $\rho$ is the distance between the origin and the closest point on the line, and $\theta$ is the angle formed by the new line and the horizontal axis.

\begin{figure}[ht!]
  \centering
  \includegraphics[width=0.8\columnwidth]{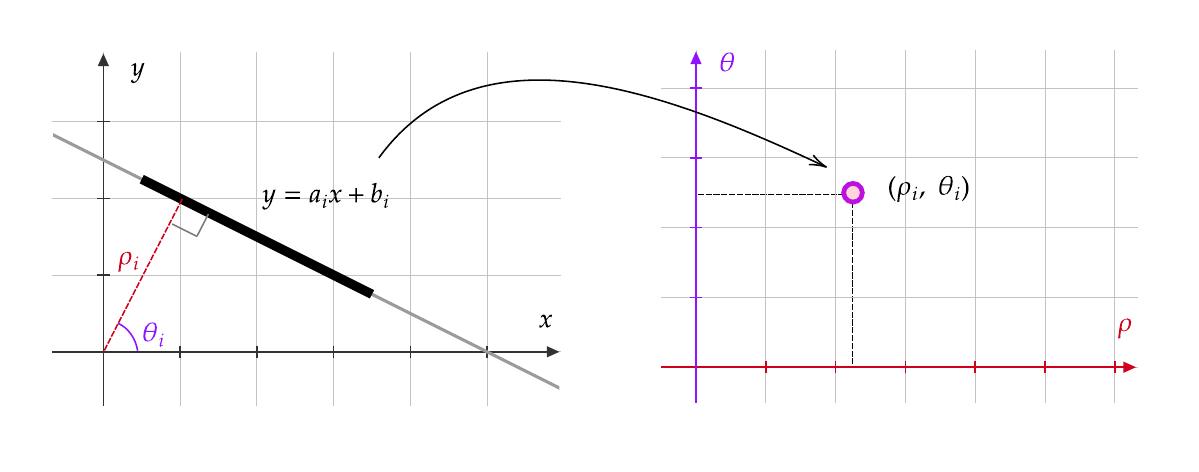}
  \caption{Example of Hough transform}
  \label{fig:hough_transform}
\end{figure}

Essentially, the Hough transform uses a point in the polar coordinate system to represent a line in the Cartesian coordinate system. In this paper, for convenience, we denote this polar coordinate system as Hough space. In Figure \ref{fig:hough_transform}, we illustrate the Hough transformation.

To detect multiple lines in an image, we first discretize the Hough space, where each point represented by $\rho$ and $\theta$ correspond to a possible line in an image pixel space. Next, by comparing the distance between the mask pixels in the image with all the lines, only lines that are close to a sufficient amount of masked pixels are selected, and the corresponding coordinates in the Hough space are also kept.

Finally, the resulting two-dimensional point features in the Hough space are then used to construct our new loss function for improving contrail segmentation.

\subsection{Combining Dice loss at original and Hough spaces}

\begin{figure}[ht!]
  \centering
  \includegraphics[width=\columnwidth]{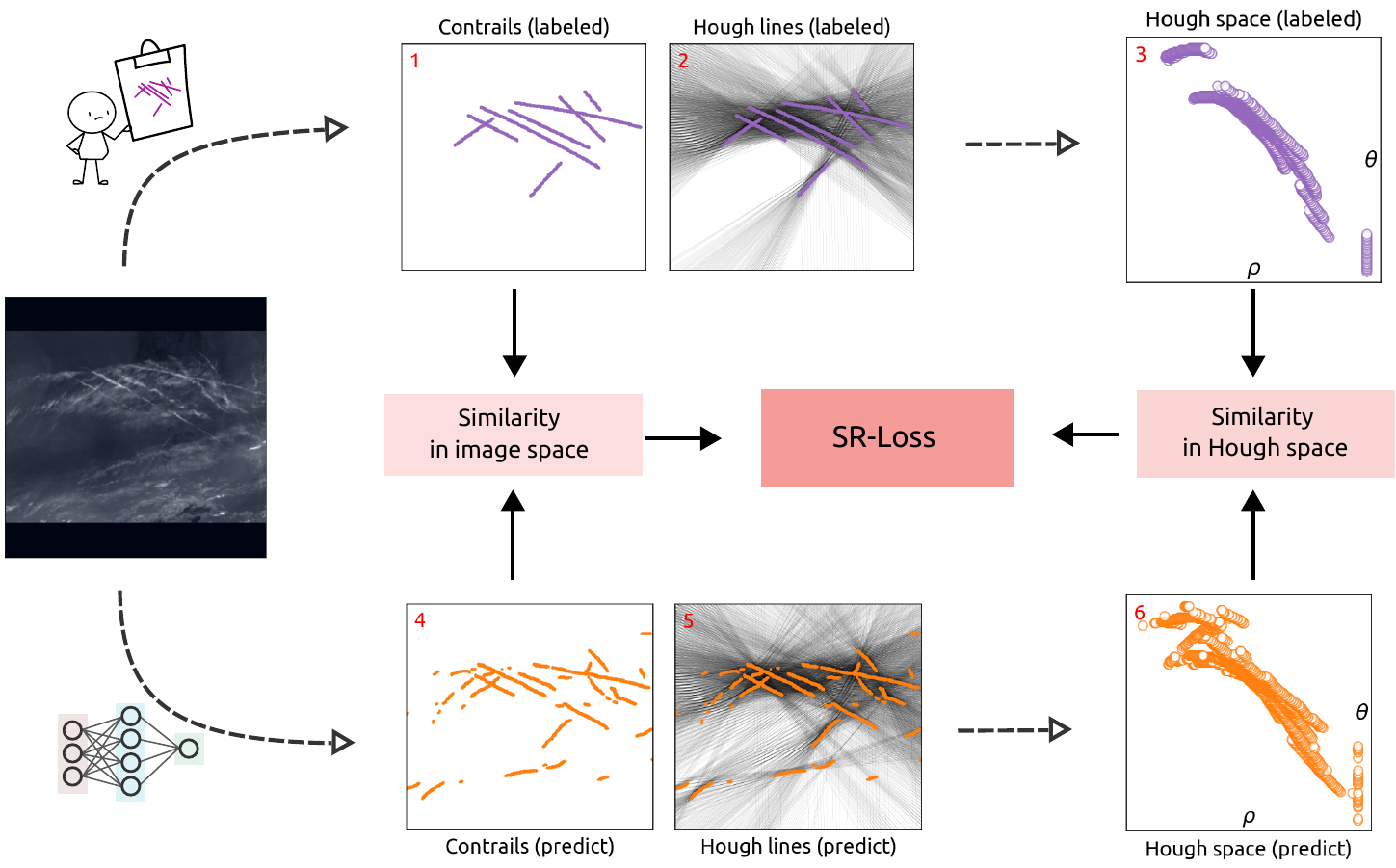}
  \caption{Customized contrail segmentation loss with Hough space transformation}
  \label{fig:hough_contrail_loss}
\end{figure}

Figure \ref{fig:hough_contrail_loss} shows an overview of how we constructed the new loss function, SR Loss, to improve the detection of contrails. In the labeled subplots of Figure \ref{fig:hough_contrail_loss}, plots 1 and 4 show the manually labeled contrail masks and predicted contrail masks from the neural network model during the training process.

Plots 2 and 5 in Figure \ref{fig:hough_contrail_loss} visualize the lines that are associated with the points in the contrail masks. We can observe that, in general, there are fewer lines detected in the labeled mask.

Plots 3 and 6 in Figure \ref{fig:hough_contrail_loss} show the representations of detected lines at the Hough space as points. In these two plots, we can observe the difference between labels and predictions more clearly. The new loss function aims to minimize the difference between plots 1 and 4, as well as plots 3 and 6.

This new loss function has two terms. The first term incorporates the aforementioned Dice loss at the pixel space, which essentially minimizes differences between the labeled and predicted contrail masks. However, the linear feature of the contrails is not explicitly considered, which will be addressed by the second term.

The second term of the loss function deals with the similarity at the Hough space. Essentially, we minimize the difference between linear features in labeled and training image masks. 

Finally, the new loss function can be formulated as follows:

\begin{equation} \label{eq:contrail_loss}
  L_\text{SR}(p, g) = \alpha L_\text{Dice}(p, g) + (1-\alpha) L_\text{Dice}(p_h, g_h)
\end{equation}

\noindent where $p$ represents the prediction for a pixel belonging to a contrail and $g$ is the ground truth from the labeled contrail mask. $p_h$ and $g_h$ are predictions and ground truth at the Hough space. $\alpha$ is an adjustable hyperparameter that can be declared to increase or decrease the weight of the loss at two different spaces.

\section{Augmented transfer learning}

\subsection{Image augmentation}

One way in improving contrail detection is to train neural network model with a sufficiently large quantity of remote sensing images. However, large, high-quality datasets are not always available, as labeling contrails is a time-consuming task.

Image augmentation provides an efficient way to generate training data using a small labeled dataset. Essentially, this can generate several order of magnitude more images for model training based on a limited number of manually labelled images.

We apply a set of transformations, also known as image augmentations \citep{buslaev2020albumentations} to the image dataset. This way, we can create a large range of scenarios with a small number of manually labeled images, which includes:

\begin{itemize}
\item Locations of contrails: This results in different locations, orientations, and perspective of the contrails in the image frame.
\item Lighting variations: This results in different brightness and contrast.
\item Viewing angle: This results in different perspectives of the contrails due to different viewing angles of the optical sensors on the satellites.
\end{itemize}

\begin{figure}[ht!]
  \centering
  \includegraphics[width=0.7\columnwidth]{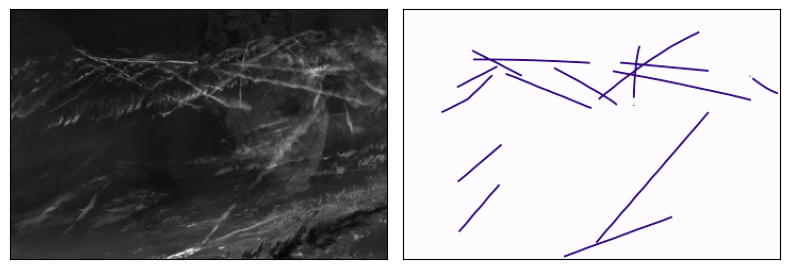}
  \caption{Original satellite image and manually created masks of contrails}
  \label{fig:augmentation_base}
\end{figure}

During the training of the model, at each step, we apply a sequence of random augmentations to an original image and its corresponding contrail mask. The base image is shown in Figure \ref{fig:augmentation_base}.

The augmentation steps essentially create a new image for the training each time an image is loaded. This process allows us to train a generalized contrail detection and segmentation model, which is robust to varying image quality and contrails.

\begin{figure}[ht!]
  \centering
  \includegraphics[width=0.8\columnwidth]{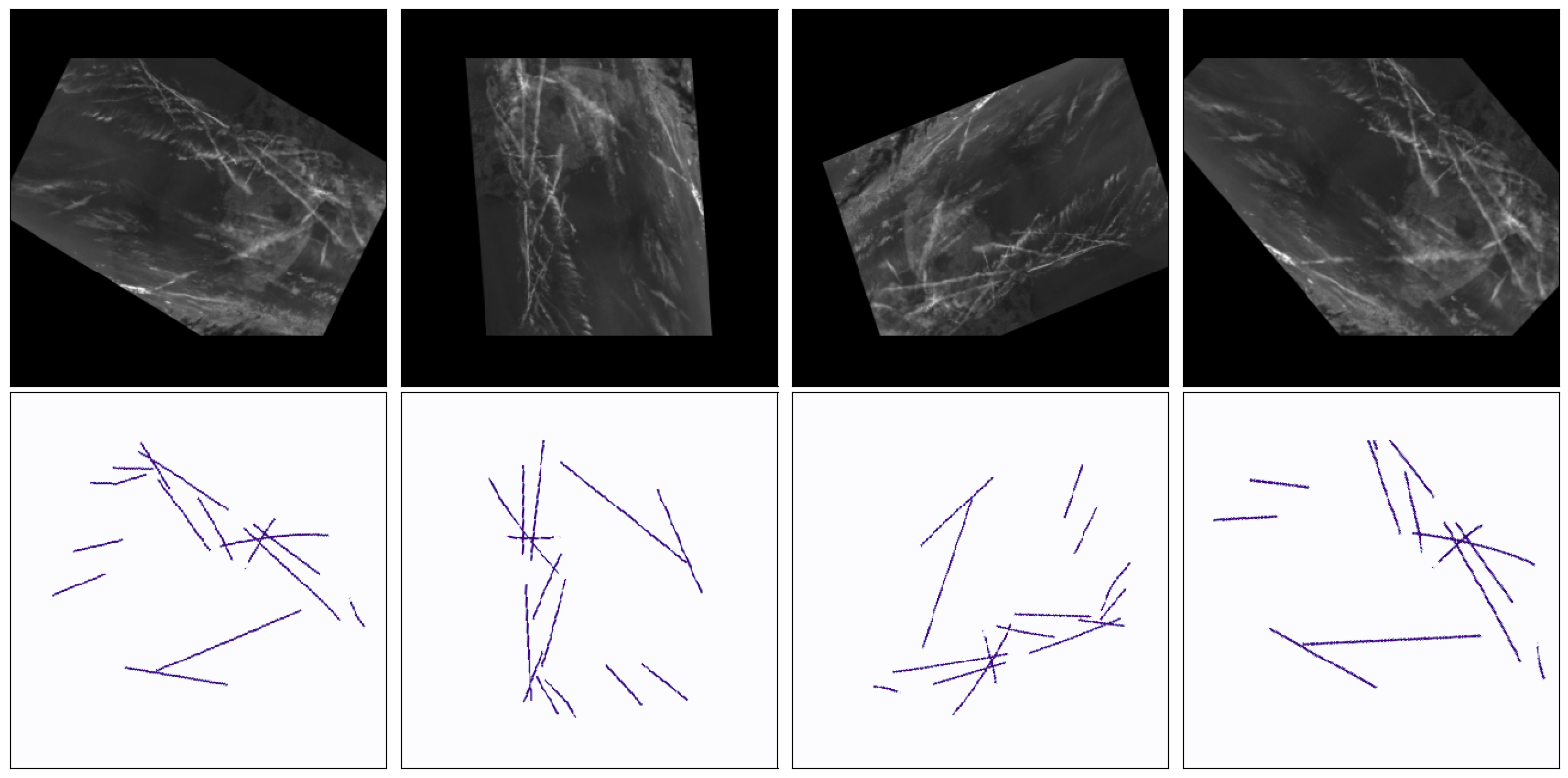}
  \caption{Random transformations applied to the original image and contrail masks. The transformations include changes of perspective, rotations, and shifts.}
  \label{fig:augmentation_transform}
\end{figure}

The first augmentation applies a transformation to the image so that the rotations, scales, positions, and perspectives of the images (and the contrail masks) are randomly altered. In Figure \ref{fig:augmentation_transform}, four different random transformations are illustrated.

In this figure, we can see that even though contrails are concentrated only in the top region of the original image, during the training, all different location scenarios are considered. As the transformed images now have different sizes, we also pad or crop the image to the same pixel size, to a multiple of 32 pixels (320 by 320 pixels in this case).

\begin{figure}[ht!]
  \centering
  \includegraphics[width=0.8\columnwidth]{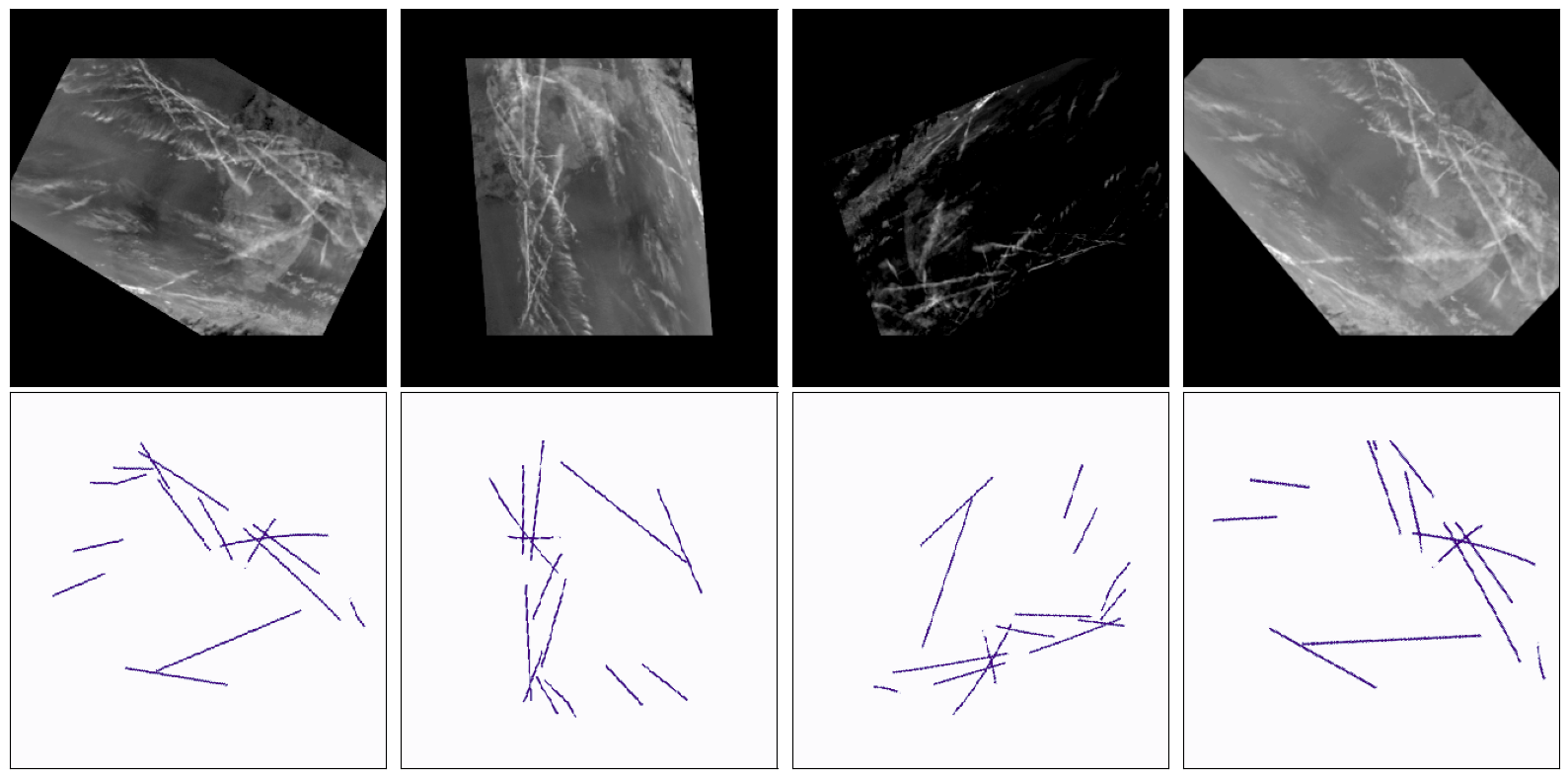}
  \caption{Random change of gamma corrections applied to the previously transformed images.}
  \label{fig:augmentation_bright_contrast_gamma}
\end{figure}

The next augmentation focuses on varying the lighting conditions of the image. For this purpose, we applied random changes in the brightness and contrast of the image. Another transformation in addition to the random brightness and contrast adjustments is the random gamma correction. This operation applies a random correction of luminance for each of the satellite images. The results for the brightness, contrast, and gamma adjustments can be seen in Figure \ref{fig:augmentation_bright_contrast_gamma}.

\subsection{Transfer learning based on pre-trained models}

The architecture also allows us to leverage the transfer learning technique, which uses a pre-trained ResNet backbone model pre-trained on a large existing image dataset, such as ImageNet \citep{deng2009imagenet}. The extracted features are then passed to the U-Net decoder, which is not pre-trained.

Transfer learning has evolved into a widely adopted strategy in contemporary machine learning methodologies. In employing transfer learning, we start with a neural network model whose weights are pre-trained with a large, generic dataset, such as the ImageNet dataset. Subsequently, the model undergoes further training with a domain-specific dataset; in our case, a manually labeled contrail dataset. The primary advantage of this approach lies in its capacity to significantly reduce the time and resources needed for model training.

The entire model is then trained with domain-specific data. In this study, a set of hand-labeled contrail images is used to train the model.

\section{Experiments and results} \label{sec:experiment}

\subsection{Evaluating the model with unseen GOES data}

The first experiment is to validate the model on unseen satellite images used for training and validation. Unlike common image objective identification tasks, where the performance can be evaluated using an easily computed ground truth, the evaluation of contrail detections is complex. Firstly, there are often multiple contrails existing in satellite images, so the performance cannot simply be examined based on the presence or absence of contrails. Secondly, persistent contrails can disperse into cirrus clouds, which become indistinguishable from clouds in many cases.

\begin{figure}[ht!]
  \centering
  \includegraphics[width=0.7\columnwidth]{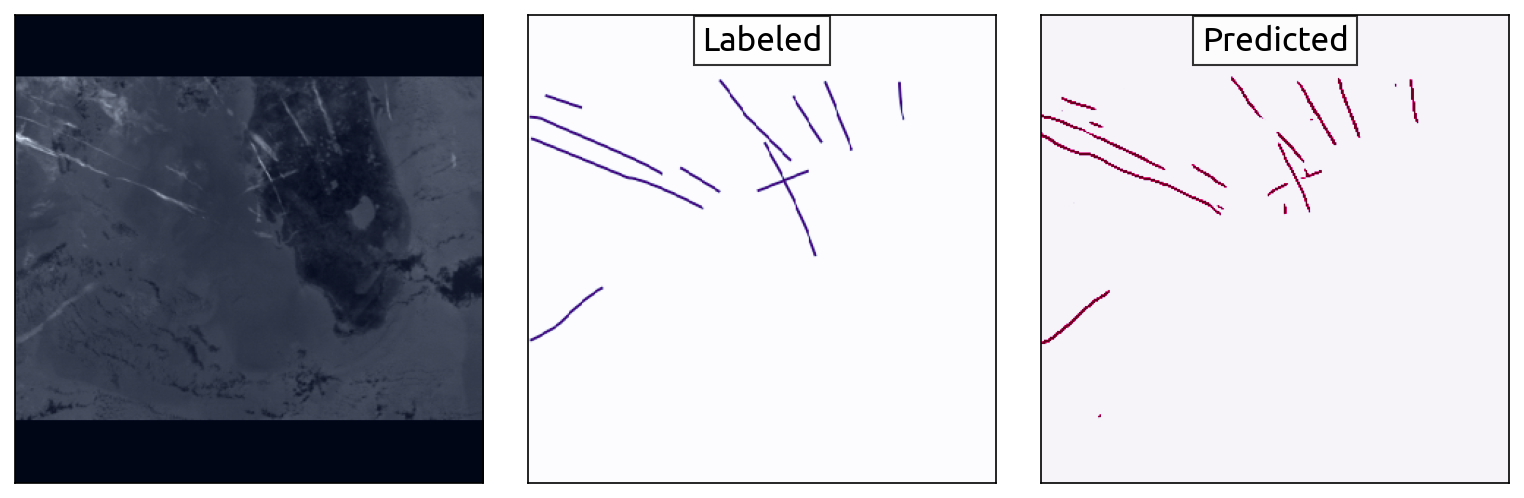}
  \includegraphics[width=0.7\columnwidth]{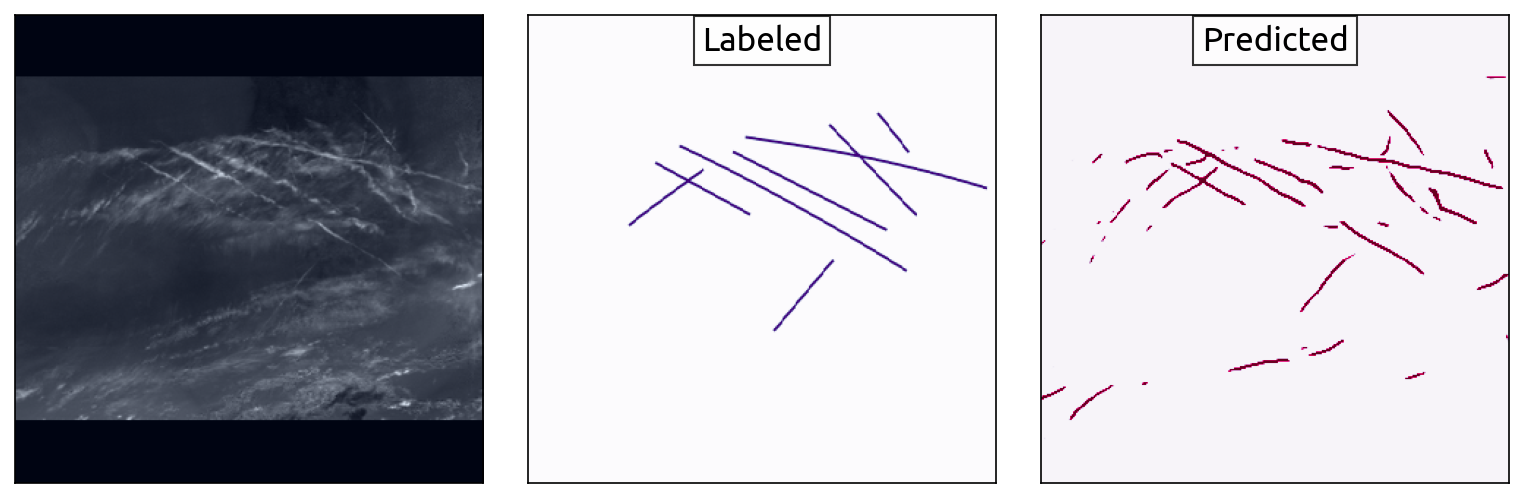}
  \includegraphics[width=0.7\columnwidth]{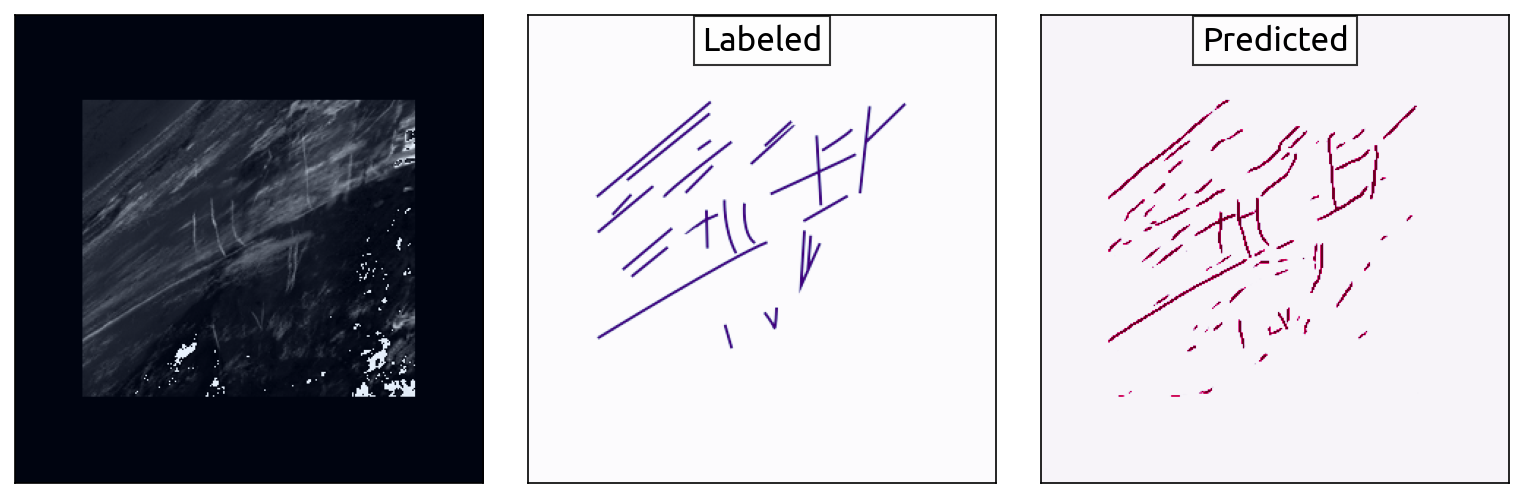}
  \caption{Application of the contrail detection model with unseen GOES images. Model trained with Dice loss.}
  \label{fig:test_example}
  \end{figure}

Thus, the performance evaluation is mostly based on the visual inspection of the results. In Figure \ref{fig:test_example}, we present the original image, manually labeled contrails, and predicted contrails using our neural network model trained with Dice loss. We can observe the following:

\begin{itemize}
  \item The first row of plots shows a relatively simple case, where the contrails are clear and newly formed. We can see that the model can detect all labeled contrails. 
  \item The second row shows a more complex situation, where newly formed contrails overlay with dispersed contrails. In this case, several contrails are not labeled; however, the neural network is still able to detect some of the segments from those unclear contrails.
  \item The third row shows the most complicated case, where the labeling is incomplete, contrails overlap with cirrus clouds, the image resolution is very low, and corrupted data occurs in the image (bottom right). In this case, we can see that the model performs worse than in the previous two examples.
\end{itemize}

\subsection{Performance under different loss functions}

Earlier, we proposed three different types of loss functions for training the ResUNet neural network model: Dice Loss, Focal Loss, and a customized SR Loss function. To evaluate the performance and influence of these loss functions, we trained different models with several thousands of steps. Models with Dice and Focal losses were trained with 8000 steps, while the model with SR loss was trained with 4000 steps, given the higher computational time required for performing Hough transformations.

\begin{figure}[ht!]
\centering
\includegraphics[width=\columnwidth]{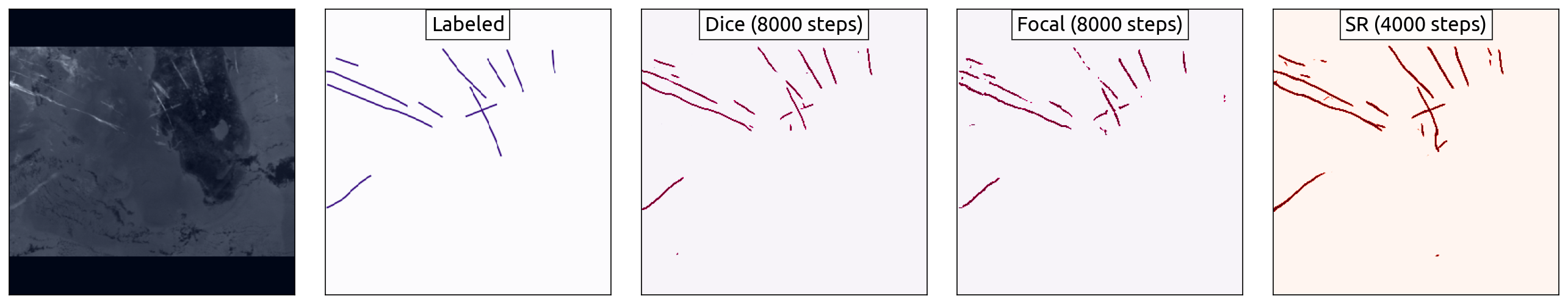}
\includegraphics[width=\columnwidth]{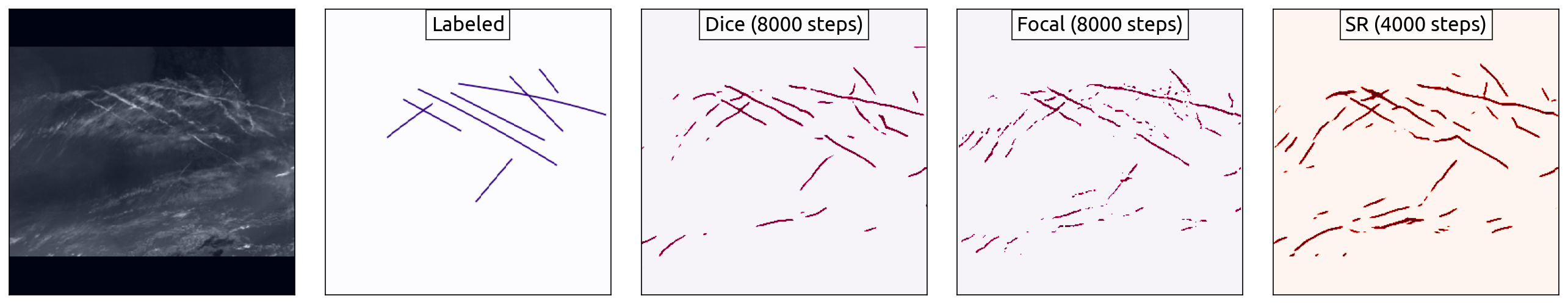}
\includegraphics[width=\columnwidth]{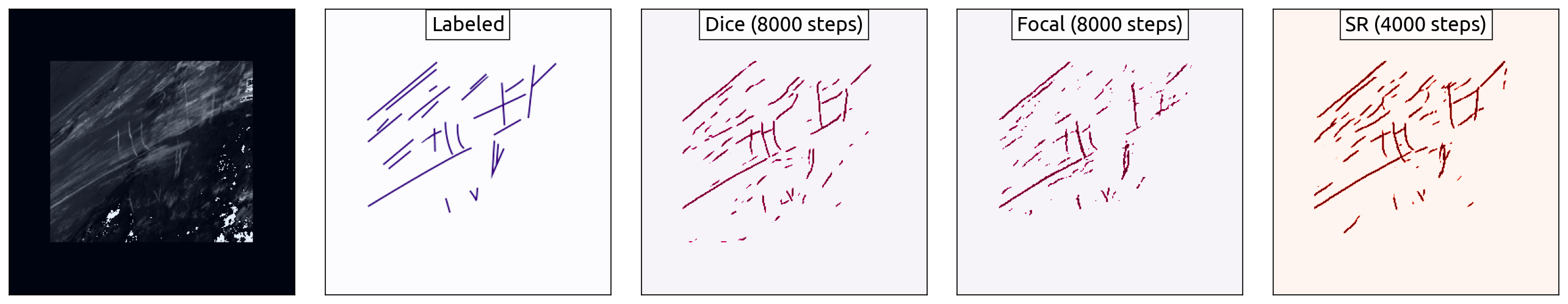}
\caption{Difference in output between Dice, Focal, and SR losses using samples from the testing dataset. Models trained with the same training dataset with 8000 steps.}
\label{fig:dice_vs_focal_vs_sr}
\end{figure}

Figure \ref{fig:dice_vs_focal_vs_sr} shows the resulting detected contrails using three models on three unseen testing images. We can see that the detection performance is similar when the situation is relatively \emph{simple}, where clear and newly formed contrails are seem with a clean background.

The detection becomes generally more challenging when the satellite image becomes more \emph{complicated}, for example, when cirrus clouds are mixed with contrails at different ages (second satellite image) or data errors occur in the image (third satellite image). In these cases, both Dice and Focal losses tend to identify scattered short segments of contrails and fail to produce parallel contrails that are very close to each other (top left of the third GOES image).

The model trained with SR loss can overcome these drawbacks, as it can focus on forming masks for longer lines of contrails, and it also performs well in segmenting contrails that are very close, which can be seen in the last two images from the previous example.

\begin{figure}[ht!]
\centering
\includegraphics[width=0.9\columnwidth]{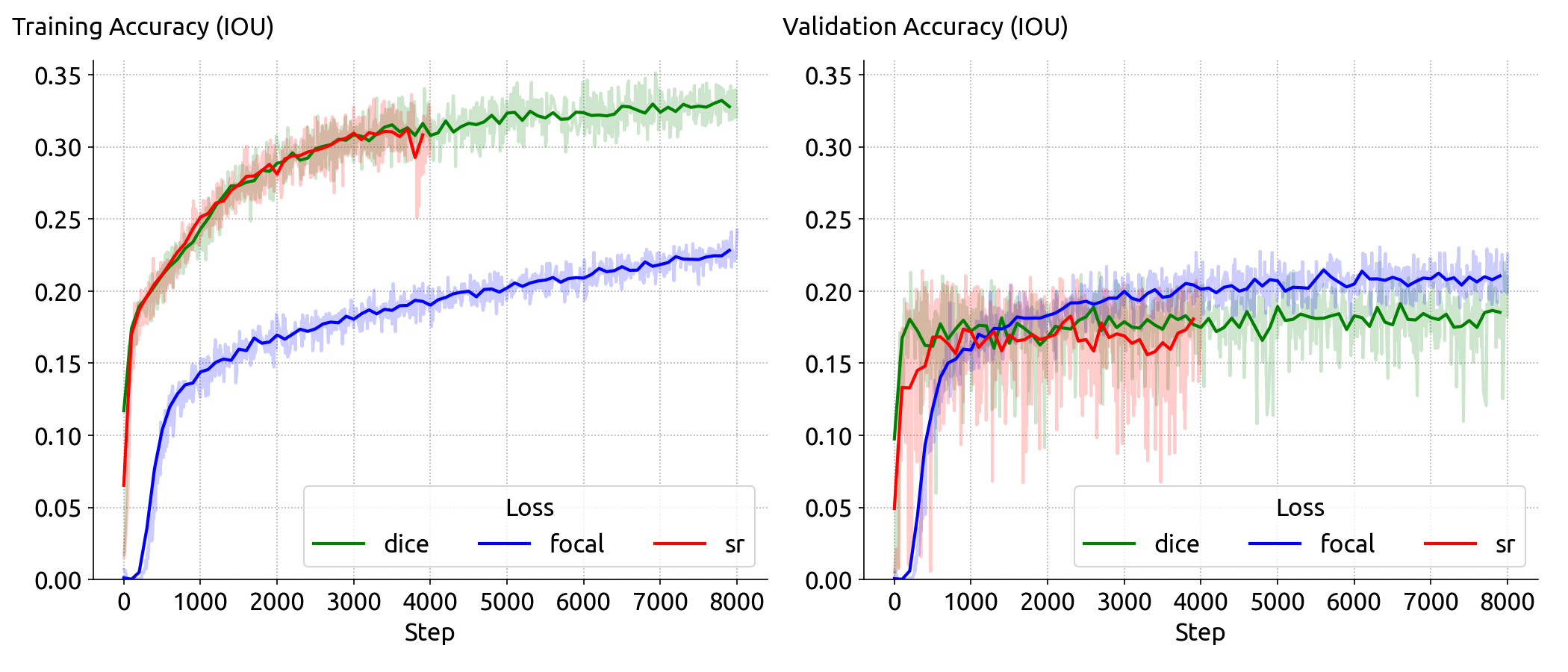}
\caption{Training and validation errors during training with different Dice and Focal loss functions.}
\label{fig:iou_dice_focal_sr}
\end{figure}

To provide more insights into the training process, Figure \ref{fig:iou_dice_focal_sr} shows the evolution of the accuracy regarding the Intersection over Union (IoU) for the detection of contrails. With our manually labeled data, the IoU metric stabilizes between 0.15 and 0.2 for all three models after 2000 training steps.

It is worth noting that the IoU metric is not strictly a performance metric for contrail detection, due to the width of contrails and incomplete labeling of all potential contrails in each image. In later discussion section, we offer more insight into this error metric and performance.

\subsection{Evaluating the model with other image sources} \label{sec:other_source}

Since we apply augmentation techniques in the training of the neural network model, the resulting contrail detection model has demonstrated an ability to work with an extensive range of images. The contrail detection model can be directly applied to different types of image sources without additional training.

\begin{figure}[ht!]
  \centering
  \includegraphics[width=0.75\columnwidth]{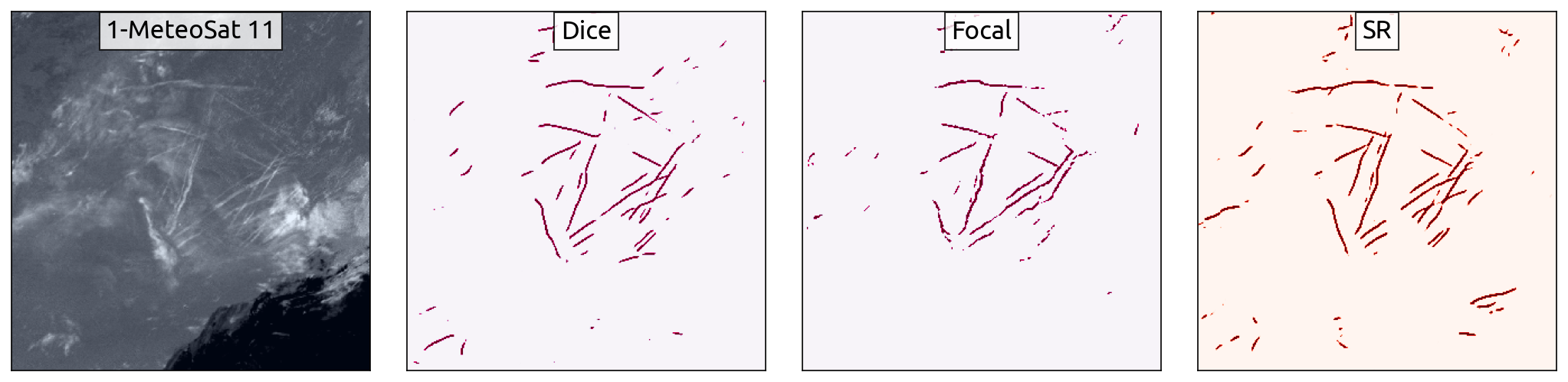}
  \includegraphics[width=0.75\columnwidth]{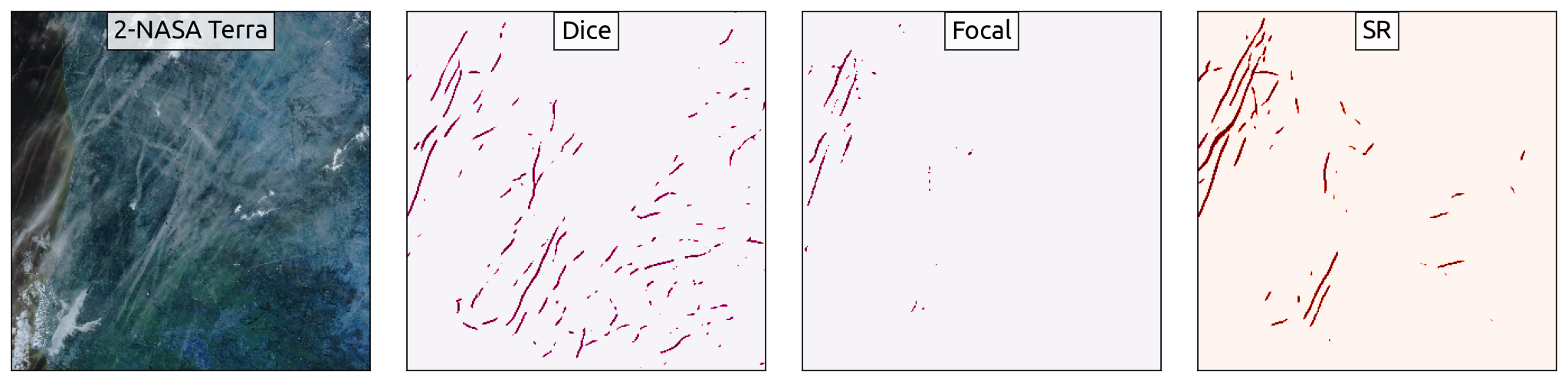}
  \includegraphics[width=0.75\columnwidth]{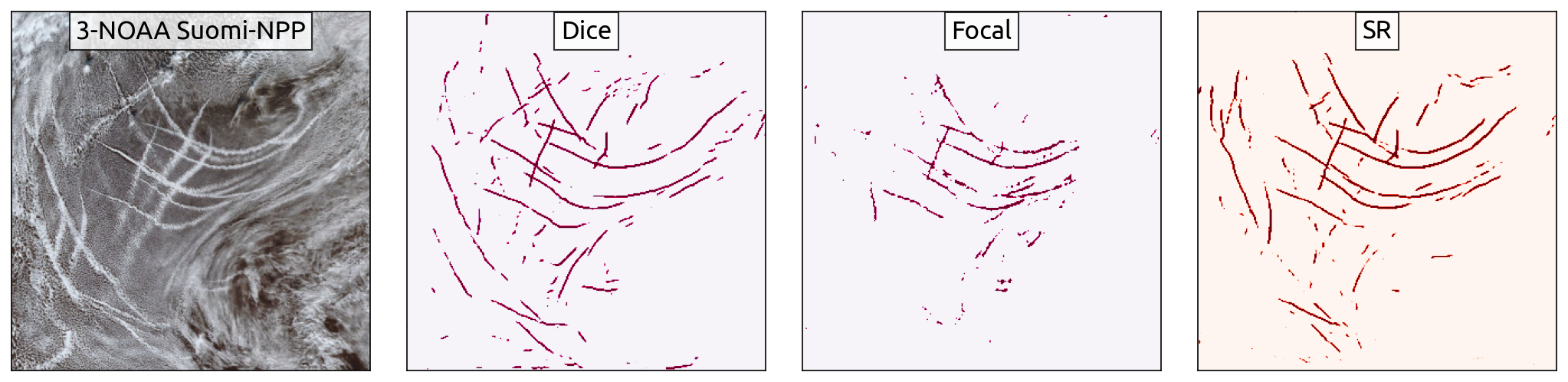}
  \includegraphics[width=0.75\columnwidth]{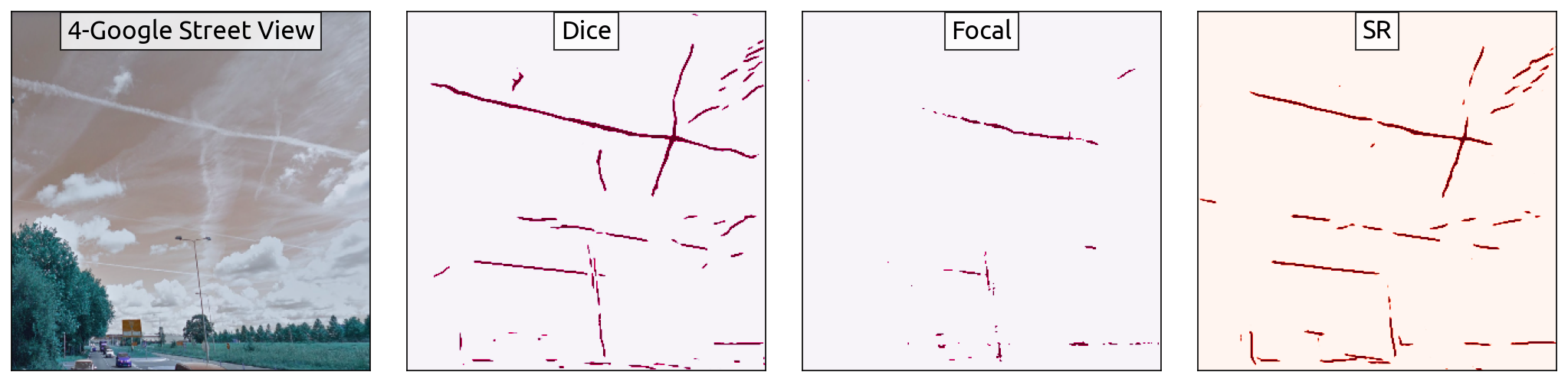}
  \caption{Contrails detected on different types of image sources. Segmentation model trained with three different losses, including the customized SR Loss function.}
  \label{fig:different_source}
\end{figure}

Figure \ref{fig:different_source} presents examples of the model's application to four distinct image sources. The first image originates from MeteoSat, which shares similar image properties with the GOES satellite imagery used during our training phase. The second image is a color photograph from the NASA Terra satellite, where the model has proven capable of managing a broad dynamic range of color inputs.

The third image in Figure \ref{fig:different_source} illustrates contrails from ships, as observed by the NOAA Suomi-NPP satellite. Despite their curved shapes, influenced by wind patterns, these ship contrails are still identifiable by our model. In the final image, we experiment with an extreme example - a random Google Street View photograph taken in the vicinity of Amsterdam's Schiphol Airport. It is clear that the model maintains consistent performance, even when applied to non-satellite images.

\section{Discussions} \label{sec:discussions}

In this section, we will further discuss the insights of the contrail segmentation approach taken in this study, focusing on uncovering details in the data, software implementation, performance, and limitations.

\subsection{Data and few-shot learning}

The challenge of generating ample training data is a well-recognized bottleneck in supervised machine learning, especially for image labeling tasks. This study addresses this issue through a two-pronged approach: leveraging few-shot learning and utilizing image augmentation.

In total, we annotated 30 satellite images for contrail segmentation. Of these, 20 images were allocated for training, and the remaining 10 were reserved for evaluation.

Unlike traditional end-to-end training that requires large datasets, the few-shot learning technique employed in this study allows for model generalization with a significantly smaller number of samples. This approach begins with a generalized segmentation model pre-trained on a large dataset and fine-tunes it specifically for contrail segmentation. Augmenting these few-shot images further amplifies the training set by several orders of magnitude, thereby enhancing the model's robustness and adaptability.

As demonstrated in Section \ref{sec:other_source}, the model exhibits exceptional performance even when applied to entirely new sources of image data. This suggests that our approach is not only effective but also scalable. With the introduction of more diverse image sources, we anticipate further improvements in model performance.

For further refinement, we recommend compiling a dataset of carefully annotated images from multiple satellites. This expanded dataset could then be subjected to our proposed image augmentation and few-shot learning strategies, likely resulting in an even more robust and generalized contrail segmentation model.

\subsection{Implementation}

We have shared the entire software implementation of the model (see Section \ref{sec:source_code}). The complete model is implemented in PyTorch \citep{paszke2019pytorch}, together with the pre-trained Segmentation Models \citep{iakubovskii2019}. The software repository includes the contrail model, training procedures, contrail detection examples, as well as the actual model weights trained with different parameters.

We also provide an alternative implementation in Tensorflow \citep{abadi2016tensorflow}, which contains only the basic modeling of the transfer learning model with image augmentations. However, the more advanced SR Loss and other fine-tuning of the performance are only available in PyTorch version.

This is also one of the first open-source machine learning-based contrail detection models. As the all source code is shared under the GNU General Public License, future researchers can make use of or contribute to the code freely.

\subsection{Loss functions}

In this paper, we evaluated three loss functions to examine the performance of the model: Dice Loss, Focal Loss, and SR Loss. The first two classical loss functions are commonly used in image segmentation tasks. The SR Loss is a novel loss function designed in this study to take advantage of the shapes of contrails. It considers the information in the Hough space and uses the similarity in Hough space between the target and prediction to improve the model training.

Based on the testing dataset, we can observe that the new SR Loss outperforms both Dice Loss and Focal Loss in complex situations, where multiple contrails and cirrus clouds are clustered together. The construction of such a loss function is one of the major scientific innovations of this paper.

We also provide a relatively fast implementation of SR Loss in PyTorch, which allows it to be computed on GPU during training. However, compared to the other two loss functions, SR Loss requires additional transformation to the Hough space, which is more computationally expensive and thus leads to slower training. However, we can observe that the model trained with 4000 steps using SR Loss is already better than the other two models trained with 8000 steps.

\subsection{Accuracy metrics}

\begin{figure}[ht!]
  \centering
  \includegraphics[width=0.9\columnwidth]{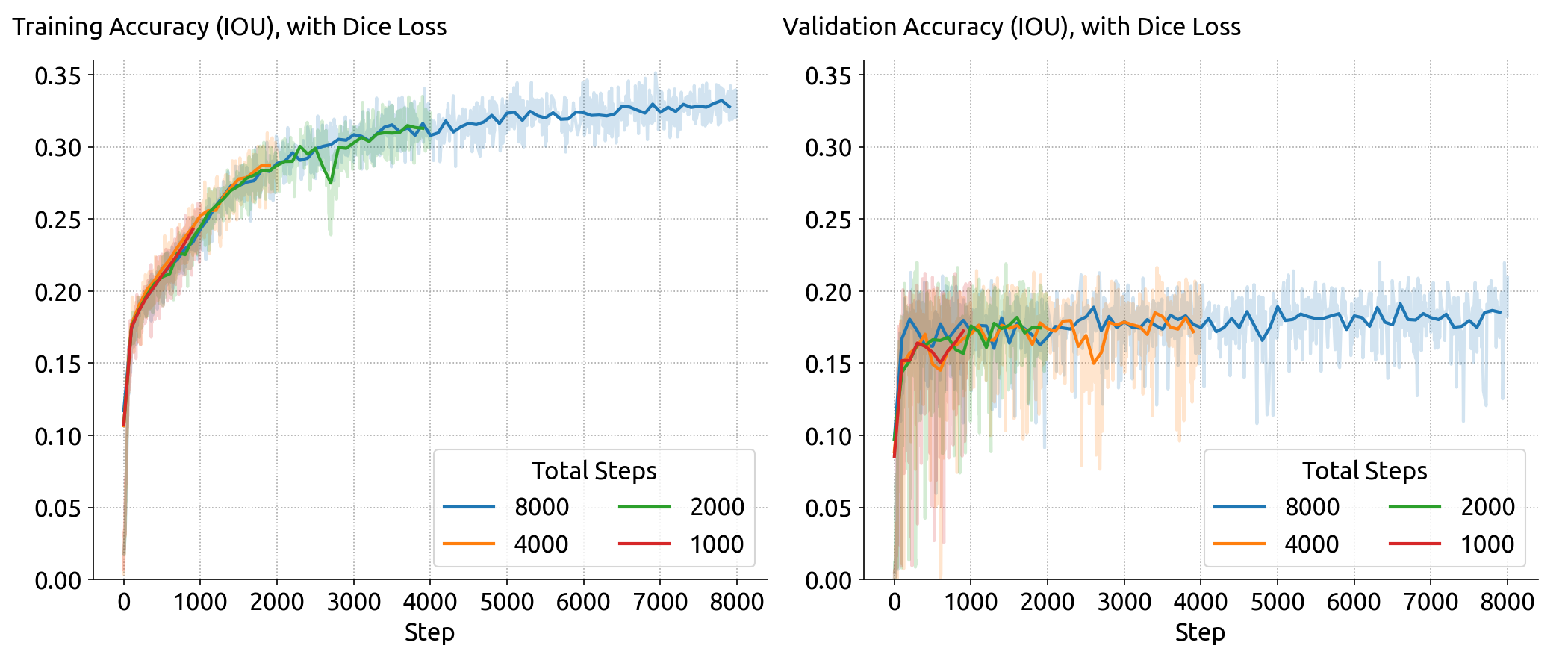}
  \caption{Training and validation errors during different rounds of training with varying numbers of steps.}
  \label{fig:iou_dice_multi_runs}
\end{figure}

In this study, we adopted the commonly used \emph{Intersection over Union} (IoU) accuracy metric to monitor the training performance. It can be seen that the validation accuracy generally stabilizes between 0.15 and 0.20 for models trained with regardless of the choice of loss functions, as shown in Figure \ref{fig:iou_dice_multi_runs}.

However, based on the visual inspection of the detected contrails, we conclude that IoU is not the best indicator for actual detection performance. This also extends to other commonly used classification metrics like F1 Score. This is due to the fact that human-labeled contrail images can not be complete, where some of the contrails or part of the contrails are not labeled.

Secondly, the contrail masks hardly ever cover the entire contrail. In our labeling process, we mask contrails with lines of similar width in pixels. Hence, the edges of the contrails are not always included in the contrail mask, as shown in Figure \ref{fig:labeling}.

All these factors limit the validity of using machine learning accuracy metrics to directly judge the contrail detection performance. To examine the detection of contrails, we still have to rely visual inspections to confirm the model's performance.

\section{Conclusions} \label{sec:conclusion}

This study presents a new approach to detecting flight contrails in remote sensing imagery data using by fine-tune a pre-train U-Net segmentation model and a novel SR Loss function. This approach has proven to be highly effective in handling varied image quality, including contrast, distortions, and lighting conditions.

Our contrail detection model is trained on a very small set of carefully labeled images, in combination with image augmentation techniques. This generates diverse transformations and qualities of satellite images. Despite the limited amount of training data, the model demonstrates strong performance with completely new image sources, and it would be capable of further improvement with a richer dataset from different satellites.

Moreover, the creation of the SR Loss function, specifically designed to leverage the linear nature of contrails in the Hough space, represents a significant contribution of this study. This new loss function has demonstrated superior performance compared to traditional Dice Loss and Focal Loss functions, particularly in complex situations involving multiple contrails and clustered cirrus clouds.

The software implementation for this model, including the contrail model, training procedures, contrail detection examples, and the actual model weights trained with different parameters, has been open-sourced under the GNU General Public License, allowing future researchers to freely use or contribute to the model.

Overall, this study has opened new avenues for contrail detection using machine learning, providing an innovative solution to the lack of large hand-labeled datasets and providing potential paths for future research and model enhancement.

\section{Code and data availability} \label{sec:source_code}

The source code and data are available at: \url{https://github.com/junzis/contrail-net}. More examples of the resulting contrail detection can also be found in this repository.

\section*{Acknowledgments}

The project is supported by TU Delft Climate Action Seed Fund Grants. We would also like to thank our students Jakub Jackiewicz, Lodewijk Bakker, Michal Jackiewicz, and Olivier Heukelom for their contributions to the research during their TU Delft Capstone AI project.

\section*{Author contribution}

\begin{itemize}
  \item Junzi Sun: Conceptualization, Funding Acquisition, Methodology, Software, Writing- Original draft, Data curation, Writing - Original draft, Writing – Review \& Editing Visualization, Data Curation
  \item Esther Roosenbrand: Conceptualization, Methodology, Writing – Review \& Editing, Data Curation
\end{itemize}


\bibliography{reference}

\begin{thebibliography}{23}
\providecommand{\natexlab}[1]{#1}
\providecommand{\url}[1]{{\tt #1}}
\providecommand{\urlprefix}{URL }
\expandafter\ifx\csname urlstyle\endcsname\relax
  \providecommand{\doi}[1]{https://doi.org/\discretionary{}{}{}#1}\else
  \providecommand{\doi}{https://doi.org/\discretionary{}{}{}\begingroup
  \urlstyle{rm}\Url}\fi

\bibitem[{Abadi et~al.(2016)Abadi, Agarwal, Barham, Brevdo, Chen, Citro,
  Corrado, Davis, Dean, Devin et~al.}]{abadi2016tensorflow}
Abadi, M., Agarwal, A., Barham, P., Brevdo, E., Chen, Z., Citro, C., Corrado,
  G.~S., Davis, A., Dean, J., Devin, M., et~al.: Tensorflow: Large-scale
  machine learning on heterogeneous distributed systems, arXiv preprint
  arXiv:1603.04467, 2016.

\bibitem[{Ackerman(1996)}]{ackerman1996global}
Ackerman, S.~A.: Global satellite observations of negative brightness
  temperature differences between 11 and 6.7 $\mu$m, Journal of the atmospheric
  sciences, 53, 2803--2812, 1996.

\bibitem[{Bedka et~al.(2013)Bedka, Minnis, Duda, Chee, and
  Palikonda}]{bedka2013properties}
Bedka, S.~T., Minnis, P., Duda, D.~P., Chee, T.~L., and Palikonda, R.:
  Properties of linear contrails in the Northern Hemisphere derived from 2006
  Aqua MODIS observations, Geophysical Research Letters, 40, 772--777, 2013.

\bibitem[{Buslaev et~al.(2020)Buslaev, Iglovikov, Khvedchenya, Parinov,
  Druzhinin, and Kalinin}]{buslaev2020albumentations}
Buslaev, A., Iglovikov, V.~I., Khvedchenya, E., Parinov, A., Druzhinin, M., and
  Kalinin, A.~A.: Albumentations: fast and flexible image augmentations,
  Information, 11, 125, 2020.

\bibitem[{Deng et~al.(2009)Deng, Dong, Socher, Li, Li, and
  Fei-Fei}]{deng2009imagenet}
Deng, J., Dong, W., Socher, R., Li, L.-J., Li, K., and Fei-Fei, L.: Imagenet: A
  large-scale hierarchical image database, in: 2009 IEEE conference on computer
  vision and pattern recognition, pp. 248--255, Ieee, 2009.

\bibitem[{Grewe et~al.(2017)Grewe, Dahlmann, Flink, Fr{\"o}mming, Ghosh,
  Gierens, Heller, Hendricks, J{\"o}ckel, Kaufmann
  et~al.}]{grewe2017mitigating}
Grewe, V., Dahlmann, K., Flink, J., Fr{\"o}mming, C., Ghosh, R., Gierens, K.,
  Heller, R., Hendricks, J., J{\"o}ckel, P., Kaufmann, S., et~al.: Mitigating
  the climate impact from aviation: Achievements and results of the DLR WeCare
  project, Aerospace, 4, 34, 2017.

\bibitem[{He et~al.(2016)He, Zhang, Ren, and Sun}]{he2016deep}
He, K., Zhang, X., Ren, S., and Sun, J.: Deep residual learning for image
  recognition, in: Proceedings of the IEEE conference on computer vision and
  pattern recognition, pp. 770--778, 2016.

\bibitem[{Hough(1962)}]{hough1962method}
Hough, P.~V.: Method and means for recognizing complex patterns, uS Patent
  3,069,654, 1962.

\bibitem[{Iakubovskii(2019)}]{iakubovskii2019}
Iakubovskii, P.: Segmentation Models Pytorch,
  \url{https://github.com/qubvel/segmentation_models.pytorch}, 2019.

\bibitem[{Kulik(2019)}]{kulik2019satellite}
Kulik, L.: Satellite-based detection of contrails using deep learning, Master's
  thesis, Massachusetts Institute of Technology, 2019.

\bibitem[{Lin et~al.(2017)Lin, Goyal, Girshick, He, and
  Doll{\'a}r}]{lin2017focal}
Lin, T.-Y., Goyal, P., Girshick, R., He, K., and Doll{\'a}r, P.: Focal loss for
  dense object detection, in: Proceedings of the IEEE international conference
  on computer vision, pp. 2980--2988, 2017.

\bibitem[{Mannstein et~al.(1999)Mannstein, Meyer, and
  Wendling}]{mannstein1999operational}
Mannstein, H., Meyer, R., and Wendling, P.: Operational detection of contrails
  from NOAA-AVHRR-data, International Journal of Remote Sensing, 20,
  1641--1660, 1999.

\bibitem[{McCloskey et~al.(2021)McCloskey, Geraedts, Jackman, Meijer, Brand,
  Fork, Platt, Elkin, and Van~Arsdale}]{mccloskey2021human}
McCloskey, K. J.~F., Geraedts, S.~D., Jackman, B.~H., Meijer, V.~R., Brand,
  E.~W., Fork, D.~K., Platt, J., Elkin, C., and Van~Arsdale, C.~H.: A
  human-labeled Landsat contrails dataset, in: ICML workshop on Climate Change
  2021, pp. 1--6, 2021.

\bibitem[{Minnis et~al.(2011)Minnis, Sun-Mack, Young, Heck, Garber, Chen,
  Spangenberg, Arduini, Trepte, Smith et~al.}]{minnis2011ceres}
Minnis, P., Sun-Mack, S., Young, D.~F., Heck, P.~W., Garber, D.~P., Chen, Y.,
  Spangenberg, D.~A., Arduini, R.~F., Trepte, Q.~Z., Smith, W.~L., et~al.:
  CERES edition-2 cloud property retrievals using TRMM VIRS and Terra and Aqua
  MODIS data—Part I: Algorithms, IEEE Transactions on Geoscience and Remote
  Sensing, 49, 4374--4400, 2011.

\bibitem[{Minnis et~al.(2013)Minnis, Bedka, Duda, Bedka, Chee, Ayers,
  Palikonda, Spangenberg, Khlopenkov, and Boeke}]{minnis2013linear}
Minnis, P., Bedka, S.~T., Duda, D.~P., Bedka, K.~M., Chee, T., Ayers, J.~K.,
  Palikonda, R., Spangenberg, D.~A., Khlopenkov, K.~V., and Boeke, R.: Linear
  contrail and contrail cirrus properties determined from satellite data,
  Geophysical Research Letters, 40, 3220--3226, 2013.

\bibitem[{Ng et~al.(2023)Ng, McCloskey, Cui, Brand, Sarna, Goyal, Van~Arsdale,
  and Geraedts}]{ng2023opencontrails}
Ng, J. Y.-H., McCloskey, K., Cui, J., Brand, E., Sarna, A., Goyal, N.,
  Van~Arsdale, C., and Geraedts, S.: OpenContrails: Benchmarking Contrail
  Detection on GOES-16 ABI, arXiv preprint arXiv:2304.02122, 2023.

\bibitem[{Paszke et~al.(2019)Paszke, Gross, Massa, Lerer, Bradbury, Chanan,
  Killeen, Lin, Gimelshein, Antiga et~al.}]{paszke2019pytorch}
Paszke, A., Gross, S., Massa, F., Lerer, A., Bradbury, J., Chanan, G., Killeen,
  T., Lin, Z., Gimelshein, N., Antiga, L., et~al.: Pytorch: An imperative
  style, high-performance deep learning library, Advances in neural information
  processing systems, 32, 2019.

\bibitem[{Ronneberger et~al.(2015)Ronneberger, Fischer, and
  Brox}]{ronneberger2015u}
Ronneberger, O., Fischer, P., and Brox, T.: U-net: Convolutional networks for
  biomedical image segmentation, in: Medical Image Computing and
  Computer-Assisted Intervention--MICCAI 2015: 18th International Conference,
  Munich, Germany, October 5-9, 2015, Proceedings, Part III 18, pp. 234--241,
  Springer, 2015.

\bibitem[{Siddiqui(2020)}]{siddiqui2020atmospheric}
Siddiqui, N.: Atmospheric Contrail Detection with a Deep Learning Algorithm,
  Scholarly Horizons: University of Minnesota, Morris Undergraduate Journal, 7,
  5, 2020.

\bibitem[{Sudre et~al.(2017)Sudre, Li, Vercauteren, Ourselin, and
  Jorge~Cardoso}]{sudre2017generalised}
Sudre, C.~H., Li, W., Vercauteren, T., Ourselin, S., and Jorge~Cardoso, M.:
  Generalised dice overlap as a deep learning loss function for highly
  unbalanced segmentations, in: Deep Learning in Medical Image Analysis and
  Multimodal Learning for Clinical Decision Support, pp. 240--248, Springer,
  2017.

\bibitem[{Vazquez-Navarro et~al.(2010)Vazquez-Navarro, Mannstein, and
  Mayer}]{vazquez2010automatic}
Vazquez-Navarro, M., Mannstein, H., and Mayer, B.: An automatic contrail
  tracking algorithm, Atmospheric Measurement Techniques, 3, 1089--1101, 2010.

\bibitem[{Weiss et~al.(1998)Weiss, Christopher, and Welch}]{weiss1998automatic}
Weiss, J.~M., Christopher, S.~A., and Welch, R.~M.: Automatic contrail
  detection and segmentation, IEEE transactions on geoscience and remote
  sensing, 36, 1609--1619, 1998.

\bibitem[{Zhang et~al.(2017)Zhang, Shang, and Zhang}]{zhang2017verification}
Zhang, J., Shang, J., and Zhang, G.: Verification for different contrail
  parameterizations based on integrated satellite observation and ECMWF
  reanalysis data, Advances in Meteorology, 2017, 1--11, 2017.

\end{thebibliography}

\end{document}